\documentclass[runningheads]{llncs}

\usepackage[misc]{ifsym}
\usepackage{graphicx}
\usepackage{amsmath}
\usepackage{amssymb}
\usepackage{amsfonts}
\usepackage{bm}
\def\mathbi#1{\textbf{\em #1}}

\begin{document}

\title{Self-Attention Enhanced Patient Journey Understanding in Healthcare System}

\titlerunning{Self-Attention Enhanced Patient Journey Understanding}

\toctitle{Self-Attention Enhanced Patient Journey Understanding in Healthcare System}

\author{Xueping Peng\inst{1}(\Letter) \and
Guodong Long\inst{1} \and
Tao Shen\inst{1} \and
Sen Wang\inst{2} \and
Jing Jiang\inst{1}}

\tocauthor{Xueping~Peng,Guodong~Long,Tao~Shen,Sen~Wang,Jing~Jiang}
\authorrunning{X. Peng et al.}
%
\institute{AAII, FEIT, University of Technology Sydney, Australia\\
\email{\{xueping.peng, guodong.long, jing.jiang\}@uts.edu.au, Tao.Shen@student.uts.edu.au} \and
School of Information Technology and Electrical Engineering, The University of Queensland, Australia
\email{sen.wang@uq.edu.au}}

\maketitle              

\begin{abstract}
Understanding patients' journeys in healthcare system is a fundamental prepositive task for a broad range of AI-based healthcare applications. This task aims to learn an informative representation that can comprehensively encode hidden dependencies among medical events and its inner entities, and then the use of encoding outputs can greatly benefit the downstream application-driven tasks. A patient journey is a sequence of electronic health records (EHRs) over time that is organized at multiple levels: patient, visits and medical codes. The key challenge of patient journey understanding is to design an effective encoding mechanism which can properly tackle the aforementioned multi-level structured patient journey data with temporal sequential visits and a set of medical codes. This paper proposes a novel self-attention mechanism that can simultaneously capture the contextual and temporal relationships hidden in patient journeys. A multi-level self-attention network (MusaNet) is specifically designed to learn the representations of patient journeys that is used to be a long sequence of activities. 
We evaluated the efficacy of our method on two medical application tasks with real-world benchmark datasets. The results have demonstrated the proposed MusaNet produces higher-quality representations than state-of-the-art baseline methods. 
The source code is available in https://github.com/xueping/MusaNet.

\end{abstract}

\section{Introduction}\label{intro}
\begin{figure}[t]
	\setlength{\belowcaptionskip}{-15pt}
	\setlength{\abovecaptionskip}{5pt} 
	\centering
	\scalebox{0.5}{\includegraphics{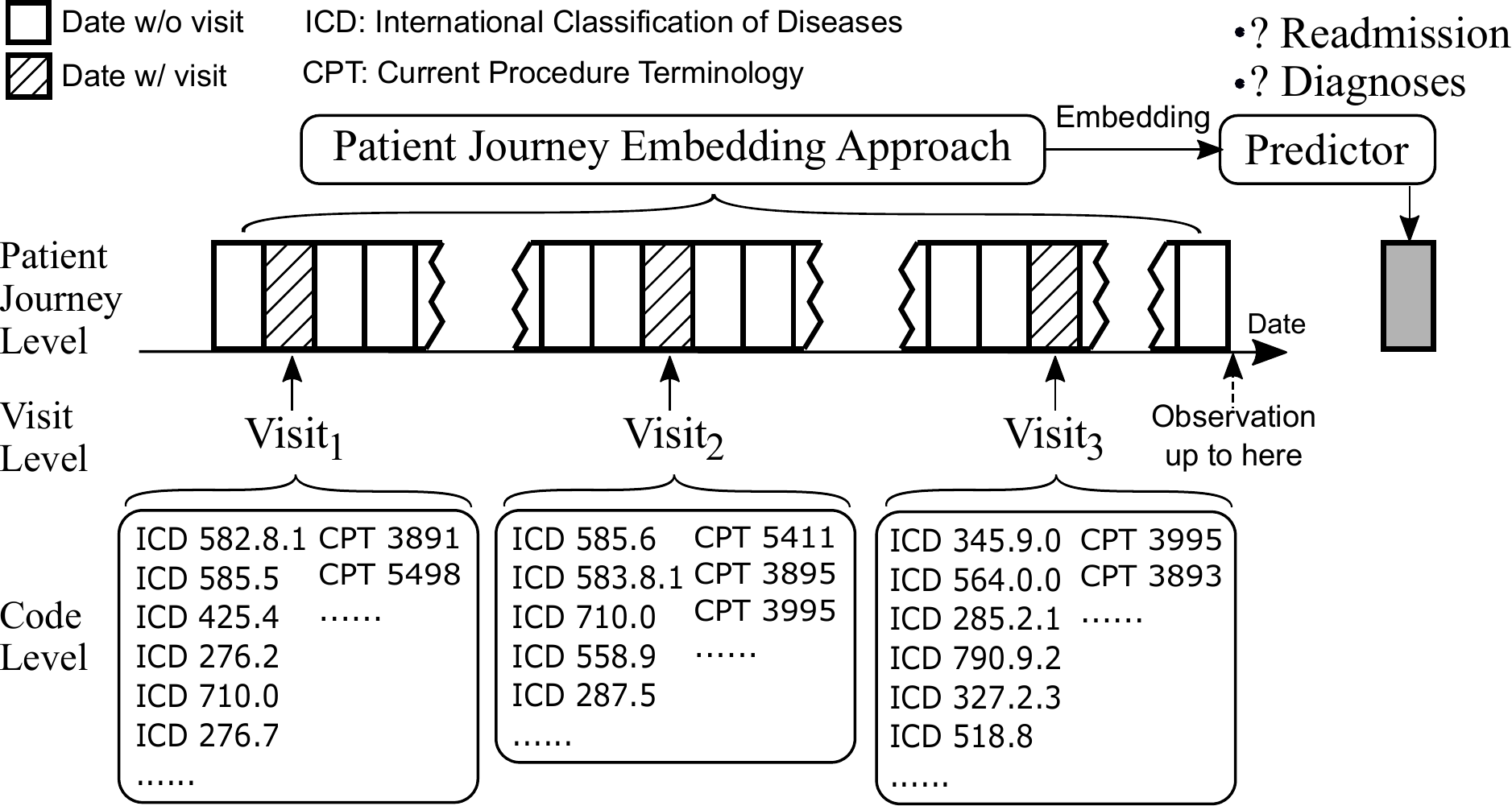}}
	\caption{\small An example of a patient's medical records in multi-level structure -- from top to bottom: patient journey level, individual visit level, and medical code level.}
	\label{ehr}
\end{figure}

Healthcare system stores huge volumes of electronic health records (EHR) that contain detailed admission information about patients over a period of time~\cite{KhoYBGGGMMMSTT19,Shickel_2018}. 
The data is organised in multi-level structure: the patient journey level, the patient visit level, and the medical code level. A useful example of this structure shown in Fig.~\ref{ehr}. In the example, an anonymous patient visits his/her doctor and is admitted to hospital on different days. The diagnoses and procedures performed at each of these visits are recorded as industry-standard medical codes. Each medical code at the lowest level records an independent observation while the set of codes at higher levels can depict the medical conditions of a patient at a given time point. At the top level, for each patient, all occurrences of medical events at different time-stamps are chained together as a patient journey, which offers more informative details. Understanding patient activity based on a patient's journey, such as re-admissions and future diagnoses, is a core research task that of significant value for healthcare system. For example, re-admission statistics can be used to measure the quality of care. Diagnoses can be used to more fully understand a patient's problems and medical research~\cite{Rajkomar_Google_2018}. However, researchers have faced many challenges in their attempts to represent patient journeys from EHR data with temporality, high-dimensionality, and irregularity~\cite{Qiao_2018}.


Recurrent neural network (RNN) has been widely used to model medical events in sequential EHR data for clinical prediction tasks~\cite{Choi_2016,Choi_Bahadori_2017,choi2018mime,ma2017dipole,Qiao_2018,Ma2018-gu}. For instance,  Ref.~\cite{choi2018mime,Choi_2016} indirectly exploit an RNN to embed the visit sequences into a patient representation by multi-level representation learning to integrate visits and medical codes based on visit sequences and the co-occurrence of medical codes. Other research works have, however, used RNNs directly to model time-ordered patient visits for predicting diagnoses~\cite{choi2016retain,Choi_Bahadori_2017,ma2017dipole,Qiao_2018,Ma2018-gu}. Convolutional Neural Network (CNN) also has been exploited to represent patient journey. For example, Ref.~\cite{nguyen2016deepr} transforms a record into a sequence of discrete elements separated by coded time gaps, and then employ CNN to detect and combine predictive local clinical motifs to stratify the risk. Ref.~\cite{song2018attend} utilises CNN in code level to learn visit embedding. These RNN and CNN based models follow ideas of processing sentences~\cite{Kim2014-pk} in documents from NLP to treat a patient journey as a document and a visit as a sentence, which only has a sequential relationship, while two arbitrary visits in one patient journey may be separated by different time intervals.
The interval between two visits, which has been largely disregarded in the existing studies on patient journey representation, can be modelled as auxiliary information fed into the supervised algorithms.

Recently, research has proposed the integration of attention mechanisms and RNNs to model sequential EHR data~\cite{choi2016retain,ma2017dipole,Rajkomar_Google_2018,Peng2019TemporalSN}. 
Ref.~\cite{vaswani2017attention} used a sole attention mechanism to construct a sequence-to-sequence model for a  neural machine translation task that achieved a state-of-the-art quality score. According to  Ref.~\cite{shen2018disan}, attention allows for more flexibility in sequence lengths than RNNs, and is more task/data-driven when modelling dependencies. Unlike sequential models, attention is easy to compute. Computation can also be significantly accelerated with distributed/parallel computing schemes. However, to the best of our knowledge, a neural network based entirely on attention has never been designed for analytics with EHR data.

To fill this gap in the literature, and address some of the open issues listed above, we propose a novel attention mechanism called masked self-attention (mSA) for temporal context fusion that uses self-attention to capture contextual information and the temporal dependencies between patient's visits. In addition, we propose an end-to-end neural network, called multi-level self-attention network (MusaNet) to understand medical outcomes using a learned representation of a patient journey based solely on the proposed attention mechanism. MusaNet constructs a multi-level self-attention representation to simultaneously represent visits and patient journeys using attention pooling and mSA layers. It is worth noting that, compared to the existing RNN and CNN-based methods, MusaNet can yield better prediction performance for a long sequence of medical records. 

Experiments conducted on two prediction tasks with two public EHR datasets demonstrate that the proposed MusaNet is superior to several state-of-the-art baseline methods. 

To summarize, our main contributions are:
\begin{itemize}
	\item a novel attention mechanism, called mSA, that uses self-attention to capture the contextual information and long-term dependencies between patient's visits;
	\item an end-to-end neural network called ``MusaNet'' that predicts medical outcomes using a learned representation of a patient journey based solely on the proposed attention mechanism; and
	\item an evaluation of the proposed model on two benchmark datasets with two prediction tasks, demonstrating that the MusaNet model is superior to all the compared methods.
\end{itemize}


\section{Background}\label{Preliminary}



\subsection{Definitions and Notations}

\begin{definition}[Medical Code] 
A medical code is a term or entry to describe a diagnosis, procedure, medication, or laboratory test administered to a patient. 
A set of medical codes is formally denoted as  $X=\{x_1, x_2,\dots, x_{|X|}\}$, where $|X|$ is the total number of unique medical codes in the EHR data.
\end{definition}

\begin{definition}[Visit]\label{visit}
A visit is a hospital stay from admission to discharge with an admission time stamp. A visit is denoted as $V_{i} = <x_i, t_i>$, where  $x_i = [x_{i1},x_{i2},...,x_{ik}]$, $i$ is the $i$-th visit, $t_i$ is the admission time of the $i$-th visit,  $k$ is the total number of medical codes in a visit.
\end{definition}


\begin{definition}[Patient Journey] 
A patient journey consists of a sequence of visits over time, denoted as $J = [V_{1},V_{2},...,V_{n}]$, where $n$ is the total number of visits for the patient. 
\end{definition}

\begin{definition}[Temporal position] \label{interval}
Temporal position refers to a difference in days between admission time $t_i$ of the $i$-th visit and admission time $t_1$ of the first visit in a patient journey, denoted as $p_{i} = |t_{i} - t_1|$, where $i = 1, \dots, n$.
\end{definition}

\begin{definition}[Task] 
Given a set of patient journeys \{$J_l\}_{l=1, \dots}$, the task is to predict medical outcomes.
\end{definition}


In the remainder of the paper, a patient's medical data means a stored and chronological sequence of {$m$} visits in a patient journey $J_i$. To reduce clutter, we omit the superscript and subscript ($i$) indicating $i$-th patient, when discussing a single patient journey.

Table~\ref{tab_notes} summarizes notations we will use throughout the paper.


\begin{table}[htbp]
\caption{ Notations for BiteNet.}\label{tab_notes}
\centering
\begin{tabular}{|c|l|}
\hline
{\bfseries Notation} & {\bfseries Description} \\
\hline
$X$      & Set of unique medical codes  \\
$|X|$     & The size of unique medical concept \\ 
$V_{i}$        & The \textit{i}-th visit of the patient\\ 
$v_{i}$        & The representation of \textit{t}-th visit of the patient\\ 
$\bm{v}$   & Sequence of $n$ visit embeddings of the patient\\ 
${x}_{i}$     & Set of medical codes in $V_{i}$  \\ 
$\bm{x}$ &  Sequence of medical codes, $[x_1, x_2, ..., x_n]$\\
${e}_{i}$ & Set of medical code embeddings in $\bm{x}_{i}$  \\ 
$\bm{e}$ &  A sequence of medical code embeddings, $[e_1, e_2, ..., e_n]$\\
${J}$       & A  patient  journey  consisting of  a  sequence  of  visits  over  time\\ 
$d$       & The dimension of medical code embedding\\ 
\hline
\end{tabular}
\end{table}

\subsection{Medical Code Embedding}

The concept of word embedding was introduced to medical analytics by Ref.~\cite{Mikolov_2013_b,Mikolov_2013} as a way to learn low-dimensional real-valued distributed vector representations of medical concepts for downstream tasks instead of using discrete medical codes. Hence, medical code embedding is a fundamental processing unit for learning EHRs in a deep neural network. Formally, given a sequence or set of medical concepts $\textit{\textbf{c}}=[c_1, c_2, ..., c_n]\in \mathbb{R}^{|C| \times n}$, where $c_i$ is a one-hot vector, and $n$ is the sequence length.
In the NLP literature, a word embedding method like word2vec \cite{Mikolov_2013_b,Mikolov_2013} converts a sequence of one-hot vectors into a sequence of low-dimensional vectors $\textit{\textbf{e}} = [e_1, e_2, ..., e_n] \in \mathbb{R}^{d \times n}$, where $d$ is the embedding dimension of $e_i$. This process can be formally written as $\textbf{\textit{e}} = W^{(e)}\textit{\textbf{c}}$, where $W^{(e)} \in \mathbb{R}^{d \times |C|}$ is the embedding weight matrix, which can be fine-tuned during the training phase. 

A visit can be represented by the set of medical codes embedded with real-valued dense vectors. A straightforward approach to learning this representation $v_i$ is to sum the embeddings of medical codes in the visit, which is written as
\begin{equation}
\label{visit_embedding}
v_i = \sum_{e_{ik} \in {e}_{i}} e_{ik},
\end{equation}
where ${e}_{i}$ is the set of medical code embeddings in the $i$-th visit, and $e_{ik}$ is the $k$-th code embedding in ${e}_{i}$. A visit can also be represented as a real-valued dense vector with a more advanced method, such as self-attention and attention pooling discussed below.

\subsection{Attention Mechanism}
\subsubsection{Vanilla Attention.}

Given a patient journey consisting of a sequence of visits  $\textit{\textbf{J}} = [v_1, v_2, ...,v_m]$ and a vector representation of a query $q \in \mathbb{R}^d$, vanilla attention~\cite{Bahdanau_2014} computes the alignment score between $q$ and each visit $v_i$ using a compatibility function  $f(v_i, q)$. A softmax function then transforms the alignment scores $\alpha \in \mathbb{R}^n$ to a probability distribution $p(z|\textit{\textbf{J}}, q)$, where \textit{z} is an indicator of which visit is important to \textit{q}. A large $p(z = i|\textit{\textbf{J}}, q)$ means that $v_i$ contributes important information to $q$. This attention process can be formalized as

\begin{equation}
\alpha = [f(v_i, q)]_{i=1}^n,
\end{equation}
\begin{equation}
p(z|\textit{\textbf{J}}, q) = softmax(\alpha).
\end{equation}
The output $s$ is the weighted average of sampling a visit according to its importance, i.e.,
\begin{equation}
s = \sum_{i=1}^n p(z = i|\textit{\textbf{J}}, q)\cdot v_i.
\end{equation}

Additive attention is a commonly-used attention mechanism~\cite{Bahdanau_2014}
, where a compatibility function $f(\cdot)$ is parameterized by a multi-layer perceptron (MLP), i.e., 
\begin{equation}
\label{add_attn}
f(v_i, q) = w^T \sigma (W^{(1)}v_i+W^{(2)}q + b^{(1)})+b,
\end{equation}
where $W^{(1)} \in \mathbb{R}^{d\times d}$, $W^{(2)} \in \mathbb{R}^{d\times d}, w \in \mathbb{R}^d$ are learnable parameters, and $\sigma(\cdot)$ is an activation function. 

Ref.~\cite{shen2018disan} recently proposed a multi-dimensional (multi-dim) attention mechanism as a way to further improve an attention module's ability to model contexts through feature-wise alignment scores. 
So, to model more subtle context and dependency relationships, the score might be large for some features but small for others. Formally, $P_{ki} \triangleq p(z_k = i|\textbf{\textit{J}}, q)$ denotes the attention probability of the \textit{i}-th visit on the \textit{k}-th feature dimension, where the attention score is calculated from a multi-dim compatibility function by replacing the weight vector \textit{w} in Eq. (\ref{add_attn}) with a weight matrix. Note that, for simplicity, we ignore the subscript \textit{k} if it does not cause any confusion. The attention result can, therefore, be written as $s=\sum^n_{i=1}P_{\cdot i}\odot v_{i}$


\subsubsection{Sophisticated Self-attention Mechanism.}\label{c2csa}
Information about the contextual and temporal relationships between two individual visits $v_i$ and $v_j$ in 
the same patient journey 
$\textbf{\textit{J}}$ is captured with a self-attention mechanism~\cite{vaswani2017attention} because a context-aware representation probably leads to better empirical performance~\cite{hu2017reinforced,vaswani2017attention,shen2018disan}. 

Specifically, the query $q$ in an attention compatibility function, such as a multi-dim compatibility function, is replaced with $v_j$ , i.e.,
\begin{equation}
\label{code2code}
f(v_i, v_j) = W^T \sigma (W^{(1)}v_i+W^{(2)}v_j + b^{(1)})+b.
\end{equation}

Similar to the probability matrix $P$ in multi-dim attention, each input individual visit $v_j$ is associated with a probability matrix $P_j$ such that $P^j_{ki} \triangleq  p(z_k = i|\textit{\textbf{J}}, v_j)$. The output representation for each $v_j$ is  
\begin{equation}
\label{s_code2code}
s_j=\sum^n_{i=1}P^j_{\cdot i}\odot v_{i}
\end{equation}

The	final	output	of	self-attention	is $\textbf{\textit{s}} = [s_1, s_2, \dots, s_n]$, each of which is a visit-context embedded representation for each individual visit.

However, a fatal defect in previous self-attention mechanisms for NLP tasks is that they are order-insensitive and therefore, cannot model temporal relationships between the visits in a patient journey. Even if position embedding \cite{vaswani2017attention} is used to alleviate this problem, it is designed for consecutive natural language words rather than patient visits with arbitrary intervals. 

\subsubsection{Attention Pooling.}\label{attn_pool}
Attention pooling~\cite{lin2017structured,liu2016learning} 
explores the importance of each individual visit within an entire patient journey given a specific task. It 
works by compressing a sequence of visit embeddings from a patient journey into a single context-aware vector representation for downstream classification or regression. Formally, {$q$} is removed from the common compatibility function, which is written as 
\begin{equation}
\label{source2code}
f(v_i) = W^T \sigma (W^{(1)}v_i + b^{(1)})+b.
\end{equation}
The multi-dim attention probability matrix $P$ is defined as $P_{ki} \triangleq  p(z_k = i|\mathbi{J})$. The final output of attention pooling, which is used as sequence encoding, is similar to the self-attention mechanism above, i.e.,

\begin{equation}
\label{s_source2code}
s=\sum^n_{i=1}P_{\cdot i}\odot v_i
\end{equation}

\begin{figure}[t]
	\centering
	\includegraphics[width=0.50\textwidth]{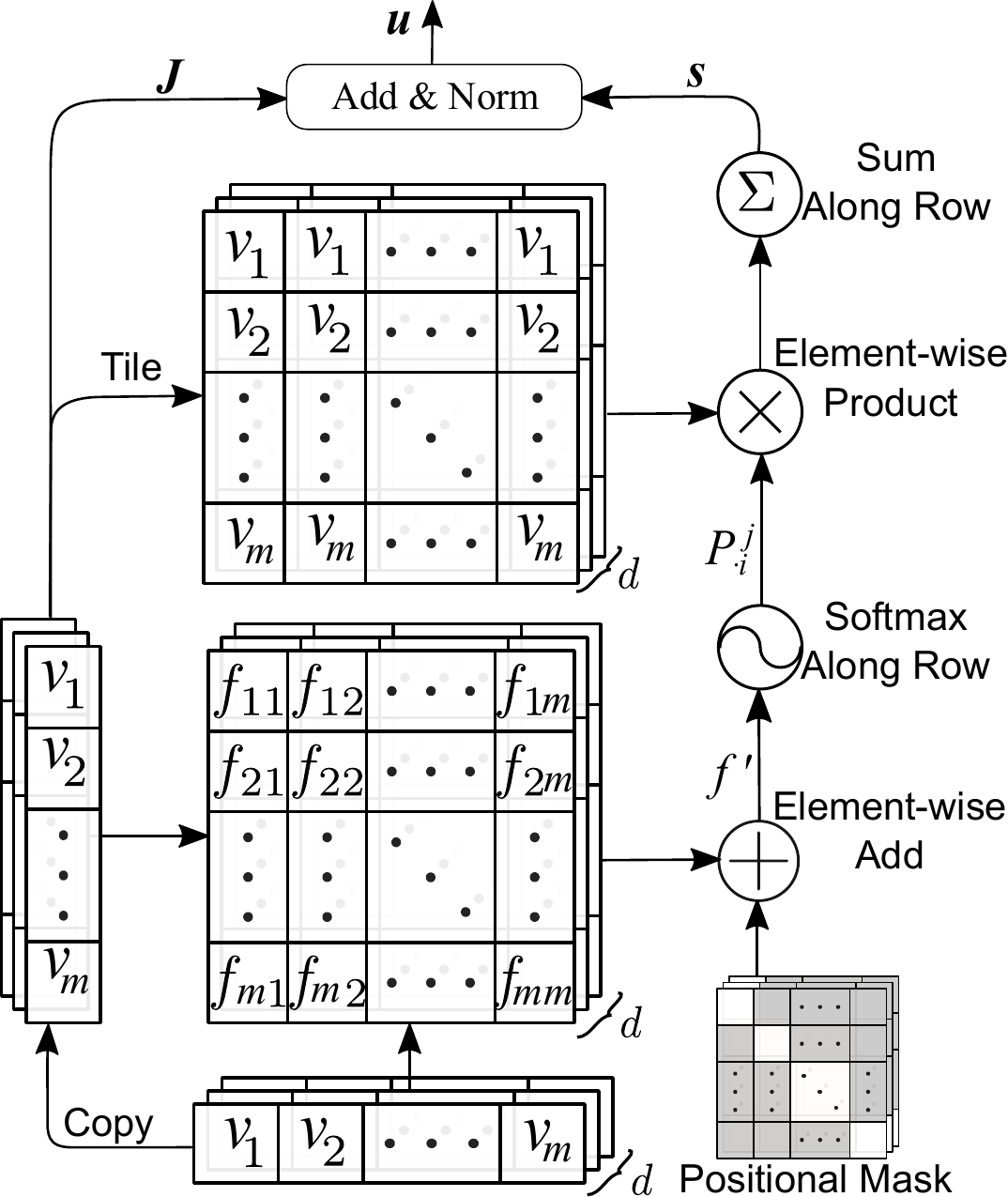}
	\caption{\small The masked self-attention mechanism (mSA). $f_{ij}$ is defined as $f(v_i, v_j)$ in Eq.~(\ref{code2code}). $f'$ is defined as $f'(v_i, v_j)$ in Eq.(\ref{temp_code2code}). $P^j_{\cdot i}$ is formally defined in Eq.(\ref{p_prob}), and \textit{\textbf{s}} is defined in Eq.~(\ref{s_code2code}).} \label{residual_self_attn}
\end{figure}

\section{Proposed Model}\label{Method}

This section begins by introducing the masked self-attention module, followed by the MusaNet context fusion module for predicting medical outcomes. 

\subsection{Masked Self-Attention}
Because self-attention was originally designed for NLP tasks, it does not consider temporal relationships (e.g., interval between visits) within inputs. Obviously, it is very important when modeling sequential medical events. Inspired by the work of Ref.~\cite{shen2018disan,shen2018bi} with masked self-attention, we developed masked self-attention (mSA) to capture the contextual and temporal relationships between visits. The structure of mSA is shown in Fig.~\ref{residual_self_attn}. 

The self-attention mechanism outlined in Eq.(\ref{code2code}) is rewritten into a temporal-dependent format: 

\begin{equation}
\label{temp_code2code}
\begin{aligned}
f'(v_i, v_j) = f(v_i, v_j) + M^{pos}_{ij},
\end{aligned}
\end{equation}

where $f(v_i, v_j) = W^T \sigma (W^{(1)}v_i+W^{(2)}v_j + b^{(1)})+b$ captures the contextual dependency between the \textit{i}-th visit and the target \textit{j}-th visit.  In this context, masks can be used to encode temporal order information into an attention output. Our approach also incorporates three masks, such as, forward $M^{fw}$, backward $M^{bw}$ and diagonal mask $M^{diag}_{ij}$, i.e., 
\begin{equation}\label{f_mask}
M^{fw}_{ij} = \begin{cases} 0,& i < j\\ -\infty, & otherwise\end{cases}
\end{equation}
\begin{equation}\label{b_mask}
M^{bw}_{ij} = \begin{cases} 0,& i > j\\ -\infty, & otherwise\end{cases}
\end{equation}
The forward mask $M^{fw}$, only attends later visits \textit{j} to earlier visits \textit{i}, and vice versa with the backward one. Here, let $M^{pos}_{ij} \in \{M^{fw}_{ij}, M^{bw}_{ij}\}$.

Given an input patient journey \textit{\textbf{J}} and a positional mask \textit{M},  $f'(v_i, v_j)$ is computed according to Eq. (\ref{temp_code2code}) followed by the standard procedure of self-attention to compute the probability matrix $P^j$ for each $v_j$ as follows:
\begin{equation}
P^j_{\cdot i} \triangleq  [p(z_k = i|\textit{\textbf{J}}, v_j)]_{k=1}^d\label{p_prob},
\end{equation}
The output \textit{\textbf{s}} is computed as Eq.(\ref{s_code2code}). Then, layer normalization (\textit{Norm}) and rectified linear unit (\textit{ReLU}) is used to generate the output \textit{\textbf{u}} as follows,
\begin{equation}
\textit{\textbf{u}} = \mathrm{Norm}(\mathrm{ReLU}(\textit{\textbf{J}} + \textit{\textbf{s}}))\label{o_output}.
\end{equation}

\begin{figure}[t]
	\centering
	\includegraphics[width=0.6\textwidth]{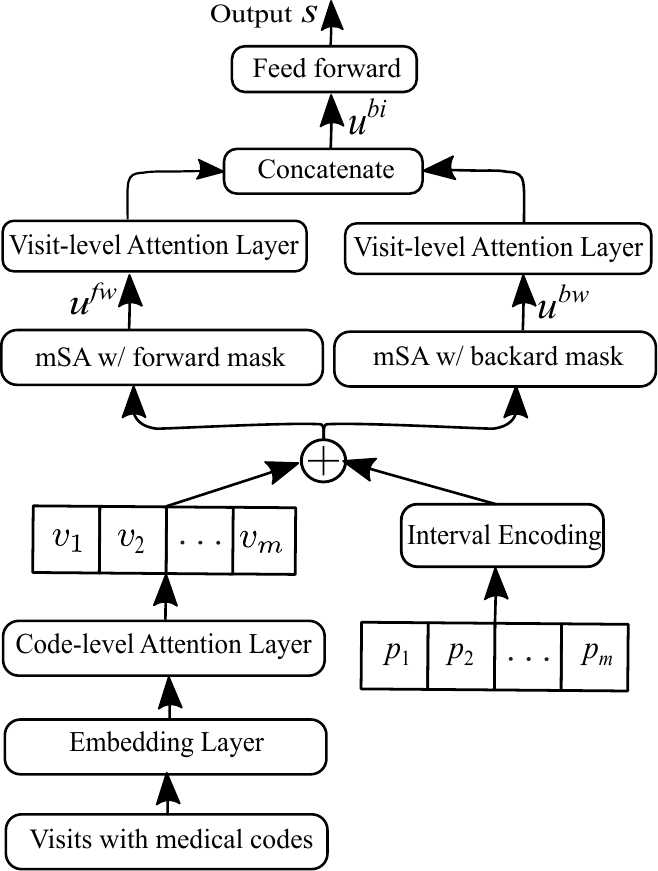}
	\caption{ \small
		The multi-level self-attention network (MusaNet) framework. The inputs are $m$ sequential visits with medical codes. 
		$m$ is the maximal number of visits in individual patient journeys over the dataset, and padding is applied if the number of visits in a patient journey is less than $m$. 
	}  \label{tesan}
\end{figure}

\subsection{Multi-level Self-attention Network}

We propose a patient journey embedding model called the ``Multi-lelvel Self-Attention Network (MusaNet)'' with mSA as its major components. The architecture of MusaNet is shown in Fig.~\ref{tesan}. 
In MusaNet, the embedding layer is applied to the input medical codes of visits, and its output is processed by 
the code-level attention layer to generate visit embeddings $[v_1, v_2, ...,v_m]$ given in Eq.(\ref{s_source2code}). The visit interval encodings are then added to the visit embeddings, which is followed by two parameter-untied mSA blocks with forward mask $M^{fw}$ in Eq.(\ref{f_mask}) and $M^{bw}$ in Eq.(\ref{b_mask}), respectively. Their outputs are denoted by $u^{fw}, u^{bw} \in \mathbb{R}^{d\times m}$. The visit-level attention layers in Eq.(\ref{s_source2code}) are applied to the outputs followed by the concatenation layer to generate output $u^{bi} \in \mathbb{R}^{2d}$. Lastly, a feed-forward layer consisting of dense and softmax layers is employed to generate a categorical distribution over targeted categories.

\subsubsection{Interval Encoding.} Although mSA incorporates information on the order of visits in a patient journey, the relative time interval between visits is an important factor in longitudinal studies. We include information on the visit intervals $p=[p_{1}, p_{2}, ...,p_{m}]$ in the sequence. In particular, we add interval encodings to the visit embeddings of the sequence. The encoding is performed by mapping interval $t$ to the same lookup table during both training and prediction. The \textit{d}-dimensional interval encoding is then added to the visit embedding with the same dimension.

\section{Experiments}\label{Experim}


\subsection{Data Description}
\subsubsection{Dataset.} MIMIC-III~\cite{Johnson_2016} is an open-source, large-scale, de-identified dataset of ICU patients and their EHRs. The dataset consists of 46k+ ICU patients with 58k+ visits collected over 11 years. In this paper, we consider two sub-datasets: Dx and Dx\&Tx, where Dx is a dataset which only includes diagnosis codes for each visit, and Dx\&Tx is another dataset which includes diagnosis and procedure codes for each visit.

\subsubsection{Data pre-processing.} We chose patients who made at least two visits. All  infrequent diagnoses codes were removed and the threshold was empirically set to 5. In summary, we extract 7,499 patients with an average of 2.66 visits per patient; the average number of diagnoses and procedures each visit are 13 and 4, respectively.

\subsection{Experiment Setup}
\subsubsection{Baseline models.} We compare the performance of our proposed model against several baseline models\footnote{GRAM~\cite{Choi_Bahadori_2017} and KAME~\cite{Ma2018-gu} and MMORE~\cite{Song2019-ol} are not baselines as they use external knowledge to learn the medical code representations.}:
\begin{itemize}
\item \textit{RNN} and \textit{BRNN}, We directly embed visit information into the vector representation $v_t$ by summation of embedded medical cods in the visit, and then feed this embedding to the GRU. The hidden state $h_t$ produced by the GRU is used to predict the (\textit{t}+1)-th visit information.
\item \textit{RETAIN}~\cite{choi2016retain}, which learns the medical concept embeddings and performs heart failure prediction via the reversed RNN with the attention mechanism.
\item \textit{Deepr}~\cite{nguyen2016deepr}: which is a multilayered architecture based on convolutional neural networks (CNNs) that learn to extract features from medical records and predict future risk.
\item \textit{Dipole}~\cite{ma2017dipole}, which uses bidirectional recurrent neural networks and three attention mechanisms (location-based, general, concatenation-based) to predict patient visit information.
 \item \textit{SAnD}~\cite{song2018attend}: which employs a masked, self-attention mechanism, and uses positional encoding and dense interpolation strategies to incorporate temporal order to generate a sequence-level prediction.
\item \textit{Transformer Encoder} (TransEnc)~\cite{vaswani2017attention}: which only considers Transformer with one encoder layer to replace mSA in MusaNet.  
\end{itemize}

\subsubsection{Validation Tasks.}
The two tasks we selected to evaluate the performance of our proposed model are to predict re-admission and future diagnosis~\cite{Rajkomar_Google_2018}. 
\begin{itemize}
	\item \textbf{Re-admission (Readm)} is a standard measure of the quality of care. We predicted unplanned admissions within 30 days following a discharge from an indexed visit. A visit is considered a “re-admission” if admission date is within thirty days after discharge of an eligible indexed visit~\cite{Rajkomar_Google_2018}.
	\item \textbf{Diagnoses} reflect the model's understanding of a patient's problems. In the experiments, we aim to predict diagnosis categories instead of the real diagnosis codes, which are the nodes in the second hierarchy of the ICD9 codes\footnote{http://www.icd9data.com}.
\end{itemize}

\subsubsection{Evaluation metrics.} 
The two evaluation metrics used are:
\begin{itemize}
\item \textbf{PR-AUC:}
(Area under Precision-Recall Curve), to evaluate the likelihood of re-admission. PR-AUC is considered to be a better measure for imbalanced data like ours~\cite{davis2006relationship}.

\item \textbf{precision@k:}
which is defined as the correct medical codes in top k divided by $\mathrm{min}(k, |y_t|)$, where $|y_t|$ is the number of category labels in the (\textit{t}+1)-th visit~\cite{Ma2018-gu}. We report the average values of precision@k in the experiments. We vary \textit{k} from 5 to 30. The greater the value, the better the performance.
\end{itemize}

\subsubsection{Implementation.}
We implement all the approaches with Tensorflow 2.0. For training models, we use RMSprop with a minbatch of 32 patients and 10 epochs. The drop-out rate is 0.1 for all the approaches. The data split ratio is 0.8:0.1:0.1 for training, validation and testing sets. In order to fairly compare the performance, we set the same embedding dimension \textit{d} = 128 for all the baselines and the proposed model.

\subsection{Results}

\setlength{\tabcolsep}{6pt}
\begin{table}[t]
	\centering
	\caption{ Prediction performance comparison of future re-admission and diagnoses (Dx is for diagnosis, and Tx is for procedure).}\label{tab_tab2}

	\begin{tabular}
	{|l|l|c|c|c|c|c|}
		\hline
		
		\textbf{Data}&\textbf{Model}&\multicolumn{1}{c|}{\textbf{Readm}}&\multicolumn{4}{c|}{\textbf{Diagnosis Precision@k}} \\
		
		\cline{4-7} & & \textbf{{\scriptsize (PR-AUC)}} &\textbf{k=5} & \textbf{k=10} & \textbf{k=20} & \textbf{k=30} \\
		\hline
		&RNN     & 0.3021 & 0.6330 & 0.5874 & 0.6977 & 0.8068 \\ 
		&BRNN    & 0.3119 & 0.6362 & 0.5925 & 0.7014 & 0.8128\\ 
		&RETAIN  & 0.3014 & 0.6498 & 0.5948 & 0.6999 & 0.8102\\ 
		&Deepr   & 0.2999 & 0.6434 & 0.5865 & 0.6981 & 0.8113\\ 
		Dx&Dipole& 0.2841 & 0.6484 & 0.5997 & 0.7034 & 0.8121\\ 
		&SAnD    & 0.2979 & 0.6179 & 0.5709 & 0.6805 & 0.7959\\ 
		&TransEnc    & 0.3116 & 0.6482 & 0.5980 & 0.7037 & 0.8139\\ 
		&MusaNet & \textbf{0.3261} & \textbf{0.6507} & \textbf{0.6069} & \textbf{0.7104} & \textbf{0.8227}\\
		\hline
		&RNN        & 0.3216 & 0.6317 & 0.5857 & 0.6973 & 0.8093 \\ 
		&BRNN       & 0.3270 & 0.6402 & 0.5961 & 0.7088 & 0.8138 \\ 
		&RETAIN     & 0.3161 & 0.6497 & 0.6021 & 0.7061 & 0.8148 \\ 
		&Deepr      & 0.3142 & 0.6391 & 0.5947 & 0.7004 & 0.8125\\ 
		Dx \& Tx&Dipole & 0.2899 & 0.6515 & 0.6097 & 0.7121 & 0.8149 \\ 
		&SAnD     & 0.2996 & 0.6242 & 0.5774 & 0.6878 & 0.8004\\ 
		&TransEnc       & 0.3137 & 0.6559 & 0.6086 & 0.7105 & 0.8144\\ 
		&MusaNet  & \textbf{0.3344} & \textbf{0.6618} & \textbf{0.6127} & \textbf{0.7146}  & \textbf{0.8242}\\ 
		\hline
	\end{tabular}

\end{table}

\subsubsection{Overall Performance.}
Tab.~\ref{tab_tab2} reports the results of the two prediction tasks on the two datasets - future re-admissions and diagnoses. The results show that MusaNet outperforms all the baselines. This demonstrates that the superiority of our framework results from the explicit consideration of inherent hierarchy of EHRs, and the contextual and temporal dependencies which are incorporated into the representations. Furthmore, we note that the performance obtained by the models remains approximately the same (RETAIN, RNN) or even drops by up to 0.43\% (Deepr) after adding the procedure to the training data over Precision@5. The underlying reason may be that the future diagnosis prediction is less sensitive to the procedures, thus the relationships among procedures cannot be well captured when predicting few diagnoses. Furthermore, by using attention mechanisms, Dipole, SAnD, TransEnc, and our MusaNet model achieve a marginal improvement when comparing the performance of using Dx\&Tx for training and that of Dx. This implies that attention could play an important role in the learned representations of the procedures.

\subsubsection{Robustness for lengthy sequence of visits.}
\begin{figure}[t]
	\centering
	\includegraphics[width=1\textwidth]{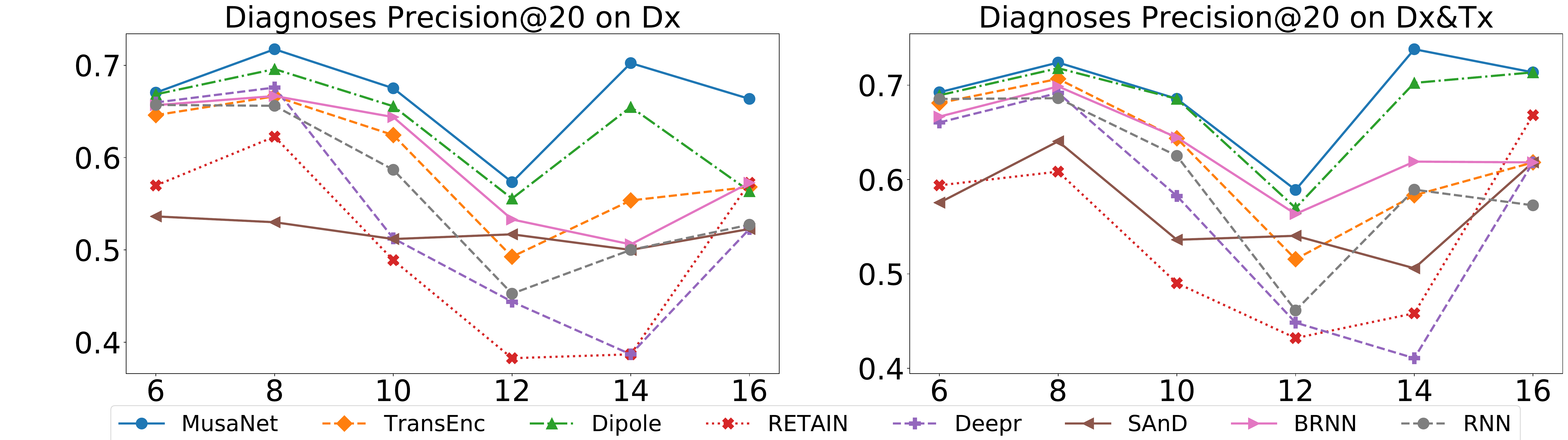}
	\caption{\small Robustness comparison regarding precision@20 on Dx and Dx\&Tx. Length of visit sequence varies from 6 to 16.} \label{fig_robustness}
\end{figure}

We conducted a set of experiments to evaluate the robustness of MusaNet by varying the length of sequential visits of a patient from 6 to 16, which are considered to be long patient journeys in the medical domain.  Fig.~\ref{fig_robustness} shows the results. Overall, the model MusaNet outperforms the other models as the length of a patient's sequential visits increases in diagnoses prediction both Dx and Dx\&Tx. In particular, the performance of Dipole is comparative to our MusaNet and follows a similar trend with an increase in the length of sequential visits. From these results, a conclusion can be drawn that the positional masks in the mSA module and interval encoding in our framework play a vital role in capturing the lengthy temporal dependencies between patient's sequential visits.

\subsubsection{Ablation Study.}
We performed a detailed ablation study to examine the contributions of each of the model's components to the prediction tasks. There are three components, which are (Attention) the two attention pooling layers to learn the visits from the embedded medical codes and learn the patient journey from the embedded visits, (PosMask) the position mask in mSA module, and (Interval) the interval encodings to be added to the learned visit embeddings. 
\begin{itemize}
	\item  \textbf{Attention:} replace each of the two attention pooling layers with a simple summation layer; 
	\item  \textbf{PosMask:} remove the forward and backward position mask in the mSA module; 
	\item  \textbf{Interval:} remove the interval encodings module; 
	\item  \textbf{MusaNet:} our proposed model in the paper.
\end{itemize}

\begin{table}[tp]
	\centering
	\caption{ Ablation Performance Comparison.}\label{tab_ablation}
	\begin{tabular}
	{|l|l|c|c|c|c|c|}
		\hline
		\textbf{Data}&\textbf{Ablation}&\multicolumn{1}{c|}{\textbf{Readm}}&\multicolumn{4}{c|}{\textbf{Diagnosis Precision@k}} \\
		
		\cline{4-7} & & \textbf{{\scriptsize (PR-AUC)}} &\textbf{k=5} & \textbf{k=10} & \textbf{k=20} & \textbf{k=30} \\
		\hline
		&Attention  & 0.3170 & 0.6371 & 0.5921 & 0.7044 & 0.8142 \\ 
		&PosMask    & 0.3033 & 0.6287 & 0.5821 & 0.7044 & 0.8125\\ 
		Dx&Interval & 0.3132 & 0.6318 & 0.5907 & 0.6977 & 0.8080\\
		&MusaNet & \textbf{0.3261} & \textbf{0.6507} & \textbf{0.6069} & \textbf{0.7104} & \textbf{0.8227}\\
		\hline
		&Attention   & 0.3278 & 0.6459 & 0.5997 & 0.7030 & 0.8104 \\ 
		Dx\&Tx&PosMask & 0.3237 & 0.6457 & 0.5957 & 0.7097 & 0.8174 \\
		&Interval  & 0.3290 & 0.6485 & 0.6050 & 0.7141 & 0.8189 \\ 
		&MusaNet  & \textbf{0.3344} & \textbf{0.6618} & \textbf{0.6127} & \textbf{0.7146}  & \textbf{0.8242}\\ 
		\hline
	\end{tabular}
	
\end{table}

		
	

From Tab.~\ref{tab_ablation}, we see that the full complement of MusaNet achieved superior accuracy to the ablated models. Specifically, we note that the position mask from the mSA module (PosMask) contributes the highest accuracy to re-admission prediction over Dx, which gives us confidence in using the position mask to learn the patient journey representations without sufficient data. Moreover, it is clear that the attention pooling component provides valuable information with the learned embeddings of the patient journey for the performance of diagnoses prediction over Dx\&Tx. Specifically, MusaNet predicted re-admissions by 2.83\% more accurately on Dx and by 1.07\% more accurately on Dx\&Tx. Similarly, MusaNet predicted diagnoses precision@5 by 2.2\% more accurately on Dx and diagnoses precision@30 by 1.4\% more accurately on Dx\&Tx.

\begin{figure}[tp]
	\centering
	\includegraphics[width=0.95\textwidth]{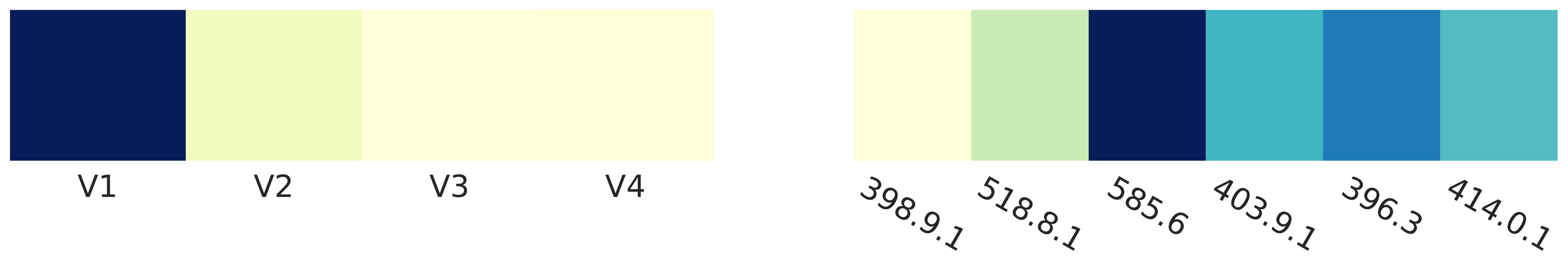}
	\caption{\small Importance of visits for patient 1 (Left), and importance of diagnoses in visit 1 (Right).} \label{fig_importance}
\end{figure}
\begin{figure}[tp]
	\centering
	\includegraphics[width=0.95\textwidth]{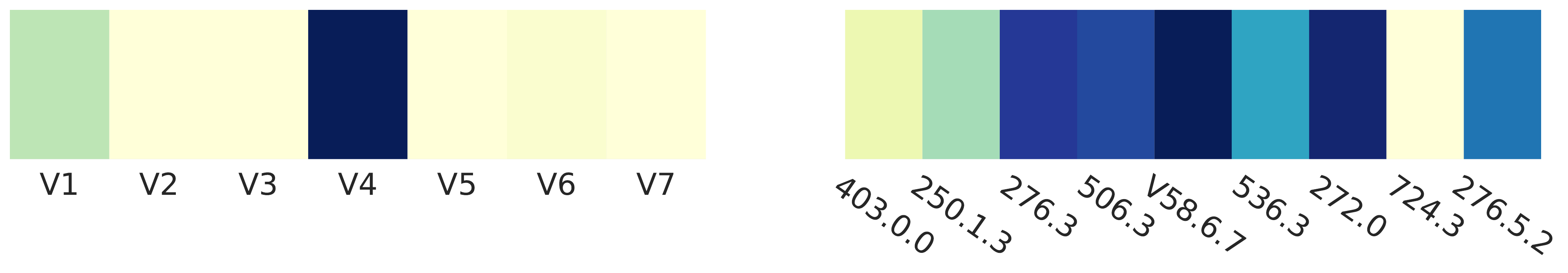}
	\caption{\small Importance of visits for patient 2 (Left), and importance of diagnoses in visit 4 (Right).} \label{fig_importance2}
\end{figure}

\subsubsection{Case Study: Visualization and Explainability.}

One aspect of this method is that it hierarchically compresses medical codes into visits and visits into patient journeys. At each level of aggregation, the model decides the importance of the lower-level entries on the upper-level entry, which makes the model explainable. To showcase this feature, we visualized two patient journeys. These examples come from the re-admission prediction task with the Dx dataset. From the importance distribution of the patient visits, we analyzed the most important visits to future re-admission. For example, Visit 1 in Fig.~\ref{fig_importance} was the most influential factor on Patient 1's re-admission. After zooming on Visit 1, we found a vital insight in Diagnosis 3 (ICD 585.6), i.e., end-stage renal disease, which would obviously cause frequent and repeated re-admissions to hospital. As shown in Fig.~\ref{fig_importance2}, Patient 2 had 7 visits. Visit 4 contributed most to the re-admissions. Again, the diagnoses reveal the cause: long-term use of insulin (ICD V58.6.7) and pure hypercholesterolemia (ICD 272.0).

\section{Related Work}\label{Relat}
\subsection{Medical Concept Embedding}
Borrowing ideas from word representation models~\cite{jha2018interpretable,Mikolov_2013,Mikolov_2013_b,wang2017learning,peng2019attentive}, researchers in the healthcare domain have recently explored the possibility of creating representations of medical concepts. Much of this research has focused on the Skip-gram model. For example, Minarro-Gimnez et al.~\cite{Minarro_2014} directly applied Skip-gram to learn representations of medical text, and Vine et al.~\cite{De_Vine_2014} did the same for UMLS medical concepts. Choi et al.~\cite{Choi_AMIA_2016} went a step further and used the Skip-gram model to learn medical concept embeddings from different data sources, including medical journals, medical claims, and clinical narratives. In other work~\cite{Choi_2016}, Choi et al. developed the Med2Vec model based on Skip-gram to learn concept-level and visit-level representations simultaneously. The shortcoming of all these models is that they view EHRs as documents in the NLP sense, which means that temporal information is ignored. 

\subsection{Patient Journey Embedding}
Applying deep learning to healthcare analytical tasks has recently attracted enormous interest in healthcare informatics. RNN has been widely used to model medical events in sequential EHR data for clinical prediction tasks~\cite{Choi_2016,Choi_Bahadori_2017,choi2018mime,ma2017dipole,Qiao_2018,Ma2018-gu}. Choi et al.~ \cite{Choi_2016,choi2018mime} indirectly exploit an RNN to embed the visit sequences into a patient representation by multi-level representation learning to integrate visits and medical codes based on visit sequences and the co-occurrence of medical codes. Other research works have, however, used RNNs directly to model time-ordered patient visits for predicting diagnoses~\cite{choi2016retain,Choi_Bahadori_2017,ma2017dipole,Qiao_2018,Ma2018-gu,Ma2018-ao,zhang2018salient}. CNN has been exploited to represent a patient journey in other way. For example, Nguyen et al.~\cite{nguyen2016deepr} transform a record into a sequence of discrete elements separated by coded time gaps, and then employ CNN to detect and combine predictive local clinical motifs to stratify the risk. 
These RNN- and CNN-based models follow ideas of processing sentences~\cite{Kim2014-pk} in documents from NLP to treat a patient journey as a document and a visit as a sentence, which only has a sequential relationship, while two arbitrary visits in one patient journey may be separated by different time intervals.
Attention-based neural networks have been exploited successfully in healthcare tasks to model sequential EHR data~\cite{choi2016retain,ma2017dipole,Rajkomar_Google_2018,song2018attend} and have been shown to improve predictive performance.

\section{Conclusion}\label{Conc}
In this paper, we proposed a novel multi-level self-attention network (\textbf{MusaNet}) to encode patient journeys. The model framework comprises a  masked self-attention module (mSA) for capturing the contextual information and the temporal relationships, attention poolings for tackling the multi-level structure of EHR data, and interval encoding for encoding temporal information of visits. As demonstrated by experiment results, MusaNet produces better representations that is validated using two downstream healthcare tasks.

\section{Acknowledgments}
This work was supported in part by the Australian Research Council (ARC) under Grant LP160100630, LP180100654 and DE190100626.

%
%
%
\bibliographystyle{splncs04}
\bibliography{bibliography}

\end{document}